\documentclass[smallextended,natbib]{svjour3}
\smartqed

\def\BibTeX{{\rm B\kern-.05em{\sc i\kern-.025em b}\kern-.08em
    T\kern-.1667em\lower.7ex\hbox{E}\kern-.125emX}}

\usepackage{url}

\usepackage{amsfonts}
\usepackage{graphicx}
\usepackage{tikz}
\usepackage{hyperref}
\hypersetup{
     colorlinks=true,
     linkcolor=blue,
     filecolor=blue,
     citecolor = blue,      
     urlcolor=cyan,
     }

\usepackage{xcolor}

\usepackage{pgfplots}
\pgfplotsset{compat=1.11}
\pgfplotsset{every axis/.append style={
                    xlabel={$x$},          
                    ylabel={$y$},          
                    label style={font=\tiny},
                    tick pos=left,
                    tick label style={font=\tiny}  
                    }}

\usepackage{todonotes}
\usepackage{paralist}

\hypersetup{
    colorlinks=true,
    linkcolor=blue,
    filecolor=magenta,      
    urlcolor=cyan,
}

\usetikzlibrary{spy}
\usetikzlibrary{calc,3d,positioning}
\usepackage{subcaption}
\usepackage[labelfont=bf,font=small]{caption}

\usepackage{xspace}

\usepackage{tikz}

\tikzstyle{edge_style} = [draw=black, line width=0.75]
\tikzstyle{edge_styleblue} = [draw=blue, line width=1]
\tikzstyle{edge_stylered} = [draw=red, line width=1]
\tikzstyle{node_style} = [circle,draw=blue,fill=blue!20!]
\tikzstyle{node_style_orange} = [circle,draw=blue,fill=orange!20!]
\tikzstyle{node_style_red} = [circle,draw=red,fill=red!20!]
\usepackage{booktabs,multirow}

\newcommand{\eg}{e.\,g.,\xspace}
\newcommand{\ie}{i.~e.,\xspace}
\newcommand{\cf}{cf.\xspace}

\usepackage{tabularx}

\begin{document}

\title{A Computational Framework for Modeling Complex Sensor Network Data Using Graph Signal Processing and Graph Neural Networks in Structural Health Monitoring}

\titlerunning{Modeling Complex Sensor Network Data Using GSP and GNNs for SHM}  
%
\author{Stefan Bloemheuvel \and Jurgen van den Hoogen \and 
Martin Atzmueller}
\authorrunning{Stefan Bloemheuvel, Jurgen v. d. Hoogen, Martin Atzmueller} 
\institute{Stefan Bloemheuvel, Jurgen van den Hoogen \at Tilburg University, Tilburg, The Netherlands, and Jheronimus Academy of Data Science, 's-Hertogenbosch, The Netherlands. \email{\{s.d.bloemheuvel,j.o.d.hoogen\}@jads.nl}
\and Martin Atzmueller \at Semantic Information Systems Group, Osnabr\"uck University, Germany \\ \email{martin.atzmueller@uni-osnabrueck.de}}

\date{}

\maketitle              

\begin{abstract}
Complex networks lend themselves to the modeling of multidimensional data, such as relational and/or temporal data. In particular, when such complex data and their inherent relationships need to be formalized, complex network modeling and its resulting graph representations enable a wide range of powerful options. In this paper, we target this -- connected to specific machine learning approaches on graphs for structural health monitoring on an analysis and predictive (maintenance) perspective. Specifically, we present a framework based on \emph{Complex Network Modeling}, integrating \emph{Graph Signal Processing} (GSP) and \emph{Graph Neural Network} (GNN) approaches. We demonstrate this framework in our targeted application domain of \emph{Structural Health Monitoring} (SHM). In particular, we focus on a prominent real-world structural health monitoring use case, \ie modeling and analyzing sensor data (strain, vibration) of a large bridge in the Netherlands. In our experiments, we show that GSP enables the identification of the most important sensors, for which we investigate a set of search and optimization approaches. Furthermore, GSP enables the detection of specific graph signal patterns (mode shapes), capturing physical functional properties of the sensors in the applied complex network. In addition, we show the efficacy of applying GNNs for strain prediction on this kind of data.


\end{abstract}

\section{Introduction}

For modeling complex data, \eg continuous sequential, multi-relational and heterogeneous data, graphs provide sophisticated means for modeling and representation. In particular, for modeling complex systems -- including those providing complex sensor data -- graphs have emerged as a natural representation. Here, \emph{Graph Signal Processing (GSP)}~\citep{stankovic2019graph} has recently emerged as a powerful analytical framework in such contexts: this is enabled both at the level of the network structure, as well as its (temporal) dynamics; GSP specifically extends on classical signal processing by providing specific analytical options on irregular structures as graphs and networks~\citep{shuman2013emerging}, which naturally accounts for irregular relations~\citep{stankovic2019understanding}.
Besides GSP, also targeted and adapted Deep Learning approaches have been adopted in complex network modeling and analysis. We specifically apply a Graph Convolutional Network~\citep{kipf2016semi,wu2020comprehensive} (GCN) approach, where one of its branches of origin is actually rooted in GSP~\cite{cheung2020graph}, the so-called spectral-based Graph Neural Networks (GNNs).

Overall, in this paper -- a substantially adapted and extended revision of~\citep{bloemheuvel2020graph} -- we present a computational framework for modeling complex sensor network data in the form of complex networks including GSP and GNN for Structural Health Monitoring (SHM)~\citep{miao2014structural,sony2019literature,abdulkarem2020wireless} and analysis. Compared to~\citep{bloemheuvel2020graph}, we have specifically extended the presentation of the proposed approach, the contextualization as well as the experimentation. Most importantly, we have included a novel component into our framework, \ie the GNN-based method for incorporating predictive analytics into our computational framework.
SHM is a multi-disciplinary field applying data-driven diagnostic methods which aim at investigating and estimating the integrity of massive complex structures.
For these, it is then the ultimate goal to increase safety, reliability, efficiency, and ultimately (cost-)effectiveness in such contexts, \eg relating to civil infrastructures such as pipeline systems, buildings, and bridges. SHM data typically includes discrete-domain signals (time series). Adapting and using insights and methods from civil engineering, signal processing, sensor technology, machine learning and data mining, \cf~\citep{miao2014structural}, the data can then be analyzed.
To the best of the authors' knowledge, as presented in this paper, this is the first time that a combination of GSP and GNNs has been applied for such a data modeling and analysis task of complex networks on real-world physical infrastructures.

Utilizing our proposed computational framework, we apply GSP and GNNs for SHM using a real-world dataset which has been collected in the context of a SHM project in the Netherlands called \emph{InfraWatch}~\citep{knobbe2010infrawatch}. In this project, data has been captured by a set of sensors installed on a major highway bridge (the so-called \emph{Hollandse Brug}), estimating the properties of traffic (\ie pressure) which is passing over the bridge. For this, sensors estimating strain, vibration, as well as ultrasonic wave sensors \citep{lynch2006summary} are typically directly attached to the respective structure.

There are two main advantages when analyzing and optimizing sensor networks \citep{capellari2018cost}. To begin with, by optimizing the sensor network with respect to sensor location and type, the number of sensors can be reduced by sampling the most optimal subset.
This leads to a cost reduction in the total SHM system. Furthermore, the amount of data that has to be analyzed is reduced significantly, speeding up the analysis. Besides that, it also increases the possibility to create real-time estimation models and reduces the data storage in the long term \citep{capellari2018cost}.
Furthermore, GSP allows for the detection of unique trends in complex data as well as the recognition of specific events. This includes, \eg the identification of traffic peaks and complex patterns found when a significant amount of pressure is applied to different sections of the bridge.
Such patterns can then indicate implicit/explicit hints and information for assessing the health of the bridge~\citep{seo2016summary}.

Both of these problems are addressed in this work, \ie how modeling and analysis are carried out and to what degree we can identify certain subsets of sensors as well as interesting patterns in the modeled complex data, respectively.
In addition, GNN methods can assist with the use case of real-time condition monitoring by forecasting the localized structural strain response of a respective monitored object. This structural strain response has recently gained an increase in attention for Condition Health Monitoring and prognosis since continuous strain measurements provide insights about the stress experienced on the bridge, in order to better characterize local weaknesses and damage to the structure compared to global responses \citep{wan2018bayesian}.
This highlights the potential of accurate forecasting of the structural stress responses by GNNs that (1) incorporate modern Deep Learning techniques and (2) complex networks to incorporate the spatial interdependence between the sensors.

%
\noindent Our contributions are outlined below:
\begin{compactenum}
    \item We suggest a theoretical framework for SHM that includes network modeling as well as complex network analysis using GSP and GNN approaches.
    \item We outline the complex network modeling and analytical methodology of the presented framework in detail, exemplified by our application use case.
    \item We use a dataset comprising real-world sensor data modeled in a complex network to illustrate the implementation of this system in a case study.
    \begin{enumerate}
    \item We present comprehensive analysis results for sensor network modeling in a resource-aware manner, with the aim of using the fewest sensors possible to recreate the provided signals.
    \item We present modeling results for signal pattern and event recognition.
    \item We show the ability of GNNs to grasp the physical nature of the sensors on a complex network embedded on a bridge.
    \end{enumerate}
\end{compactenum}

The remainder of this work is organised as follows: 
Section~\ref{sec:related} addresses related work, including an outline of essential theoretical concepts of GSP and GNN theory. Section~\ref{sec:method} then introduces our proposed structure and explains the approach in depth. Section~\ref{sec:results} introduces the case study and addresses our results. Finally, Section~\ref{sec:conclusions} concludes with an overview and interesting directions for future research.

\section{Background and Related Work}\label{sec:related}
This section discusses related work and outlines important fundamental concepts on the background of our proposed framework. We start by summarizing related work on complex networks, before we introduce the fundamental concepts of signal processing on graphs and the requisite theoretical context. For a detailed overview on GSP, we refer to \eg~\citep{ortega2018graph,stankovic2019graph}.
Next, we focus on the topic of Structural Health Monitoring.
Finally, we provide a brief summary on Graph Neural Networks, where we introduce and explain this prominent approach for Deep Learning on graphs.

\subsection{Complex Networks}

In the world of today, complex networks -- represented as graphs -- can be observed in many different areas and domains. Altogether, complex networks have proven to be an effective method for modeling structural properties in a wide range of complex systems and a number of domains, \eg~\citep{strogatz2001exploring,amaral2004complex,boccaletti2006complex,MAHS:14,Atzmueller:14:CoRR,BKA:19}.
In particular, complex structures encountered in complex (networked) systems and structures, such as computer networks, social networks, infrastructure networks, sensor networks, as well as cyber-physical networks play an important role throughout our everyday life. However, the network concept transcends such explicit structures, towards more implicit networks observed in physical structures of interdependent elements or components~\citep{bloemheuvel2020graph,worden2021towards}. In particular, in the field of complex networks and feature rich networks both the need as well as the opportunities in studying such complex network topologies, has made the use of complex network models pervasive in many fields of research such as computer science, physics, engineering and the social sciences, also joining into interdisciplinary research contexts, \cf \citep{IAGKLS:19}.

In comparison to simple homogeneous static networks, real-world networks are often dynamic and heterogeneous, with both nodes and links being represented by a collection of attributes and/or complex relationships caused by multi-relational, continuous sequential, and heterogeneous data.
Thus, the mining of so-called {\em feature-rich} networks~\citep{IAGKLS:19} is gaining increasing interest; such networks include, in particular, {\em node-attributed} and/or {\em edge-attributed} networks, where, for example, time series information obtained from sensor readings can be attached to nodes and/or edges of a network.

In this paper, we target the modeling of complex sensor network data -- regarding topological/structural dependencies and properties using complex network approaches. Specifically, we apply GSP and GNNs on the modeled networks (being represented as graphs). To the best of the authors' knowledge, this is the first time that such a combination of modeling and analysis methods has been applied for the task of SHM on real-world physical infrastructures.

\subsection{Graph Signal Processing}
Classical signal processing can be exceptionally strong in uniform, euclidean domains, \eg in the context of audio and power circuits. However, not all domains possess such a desirable feature. For instance, if examining sensors arranged along some topography of a building at distinct locations, then this arrangement will in all likelihood not resemble some kind of regular grid,  where, \eg wall and floor properties can considerably influence positioning and signal strengths of sensors. Moreover, transportation networks also resemble complex connections that are not structured uniformly. 
Some locations will serve as hubs in the network of rails, while there will be less dense connections in more urban areas. Thus, the complexity of such networks implies that the data coming from such irregular and complex structures do not lend themselves for standard tools~\citep{ortega2018graph}. This motivates, \eg including the spatial dimension towards complex modeling via GSP, extending signal processing by including irregular structures modeled as graphs~\citep{shuman2013emerging}. Signal data on a graph can then be intuitively represented as a finite set of samples, where each node contained in the graph is assigned to one sample.

\paragraph{GSP: Basic Definitions}
We define a graph as $G = (V,E)$ where $V$ are the nodes (also called vertices) and $E$ the edges. An edge $e_{ij}=(v_i,v_j)$ connects nodes $v_{i}$ and $v_{j}$, \ie they are neighbors.
The adjacency matrix $A \in \mathbb{R}^{N \times N}$ where $|V|=N$ is a square matrix such $A_{ij}=1$ if there is an edge from node $v_{i}$ to node $v_{j}$, and 0 otherwise. The number of neighbors of a node $v$ is known as the degree of $v$ and is denoted by $D_{ii} = \sum\nolimits_{j}^{}A_{ij}$.
For GSP, a graph $G$ is most often represented via the laplacian matrix $L \in \mathbb{R}^{N \times N}$,\ie the degree matrix minus the adjacency matrix; it holds several spectral properties that are desirable during GSP analysis~\citep{stankovic2019graph}. For example, the laplacian of an undirected graph is always positive semi-definite (all the eigenvalues of the matrix are non-negative). For a more detailed overview, see \citep{stankovic2019graph,ortega2018graph,ruiz2021graph}.

\begin{itemize}
    \item A graph signal is defined by associating real data values $s_{n}$ to each vertex. A graph signal is written as $s=\left[s_{0}, s_{1}, \ldots, s_{N-1}\right]^{\mathrm{T}} \in \mathbb{R}$ in vector notation.
    \item In Digital Signal Processing, a signal shift is a shift in time of length $N$, resulting in $\hat{s} = s_{n-1}$. Such an operation helps with performing autocorrelation analysis. In GSP, a signal shift is more locally defined by replacing a signal value by a combination of a neighbors signal values $V_{n}$ weighted by their respective edge weights. The two most popular graph shift operators are given by the laplacian and adjacency matrix.
    \item One of the most important transformations in classical Signal Processing is the Fourier transform, which changes the domain of a signal $x$ from the time-domain to the frequency-domain. This change of perspective makes previously difficult problems more easily solvable, since it tells you what frequencies are present in your signal and in what proportions. Translated in terms of GSP, the Graph Fourier Transform (GFT) converts the graph signal from the vertex domain into the graph spectral domain. GSP achieves this transformation via the spectral decomposition of
    \begin{equation}
        L = V \Lambda V^{-1},
    \end{equation}
    where the columns $v_{n}$ of the matrix $V$ are the eigenvectors of the laplacian $L$, and $\Lambda$ the diagonal matrix of the corresponding eigenvalues. The eigenvalues act as the frequencies on the graph~\citep{sandryhaila2014discrete}. The GFT of the signal $s$ is then calculated by $\hat{s}=U^*s$ where $U^*$ the conjugate transpose of the Fourier Basis $U$.
    \item After Graph Fourier transformation, filters can be applied. These filters transform the graph signal into the graph spectral domain. Then, unwanted frequencies are weakened or wanted frequencies are magnified by altering the Fourier coefficients. Finally, the signal is reverted to the vertex domain.
    \item Lastly, a technique to measure the smoothness of a signal on a graph is called Total Variation. Smoothness is an important subject in Graph Signal Processing since a lot of techniques depend on the assumption that nearby nodes act similar. Smoothness is expressed by the function:
    \begin{equation}
    \mathrm{TV}_{\mathrm{G}}(\mathbf{X})=\|\mathbf{X}-\mathbf{A} \mathbf{X}\|_{1}
    \end{equation}
    \noindent where $A$ is the shift operator matrix of the graph, $AX$ the shifted version of the signal and $\|\|_{1}$ the $l_{1}$-norm. In other words, it is the cumulative difference between the original signal at each node and its neighbors. One could then use the end result as a global measure for the entire signal, or also investigate the individual values for each sensor.
\end{itemize}

\subsection{Structural Health Monitoring}
The collapse of the Polcevera Viaduct in Genoa, \eg showed that good designs alone are insufficient to ensure long-term viability of civil infrastructure \citep{clemente2020monitoring}. Such structures should be continually checked to identify damage and defects and to schedule timely maintenance programs. The field of applying such data-driven diagnostics that investigate and estimate the integrity of massive structures is called Structural Health Monitoring (SHM). 

In principle, the main assumption of SHM is that global parameters (\eg mode shapes, natural frequencies) are functions of physical properties such as mass, damping, and stiffness~\citep{seo2016summary,cornwell1999environmental}.
The deformation that a part will exhibit when vibrating at its natural frequency is referred to as its mode shape.
From a Signal Processing view, mode shapes are patterns where signals and their frequencies are partitioned into different modal categories, \eg using strain and/or vibration sensors.
Both local and global characteristics can be extracted.
Specific local abnormalities of sensor data, for example, can suggest inaccurate sensor readings, motivating sensor replacement and/or maintenance. 
Global characteristics, however, could assess changes in the overall stiffness of a structure~\citep{seo2016summary}, and determine the current and future  structural capacity of a bridge~\citep{seo2016summary}.

A specific problem connecting GSP and SHM which we also target in this paper, is resource-aware optimization; via identifying the minimal subset of sensors which is required to reconstruct the signal using GSP~\citep{capellari2018cost}, the needed number of sensors for reliably capturing (sensor) data from a specific complex system (\eg a bridge) can be minimized, \eg for a sensor network monitoring dynamic/structural properties.

\subsection{Graph Neural Networks}
Besides GSP, Graph Neural Networks (GNNs) have emerged as another successful technique for modeling complex graph-structured data. One of the branches of origins in GNNs called spectral-based GNNs even originates from the GSP literature. Older efforts to build GNNs mostly consist of spatial methods that look at the neighborhood of nodes to perform message passing between pairs of nodes to agglomerate them.
This gap (spatial/spectral) has been bridged by the Graph Convolutional Network (GCN) \citep{kipf2016semi}. Since then, spatial-based techniques have developed rapidly due to their efficiency and generality~\citep{wu2020comprehensive}, \eg Graph Convolutional Networks, Graph Autoencoders and Spatial-Temporal Graph Neural Networks; here, Graph Convolutional Networks gained most attention \citep{wu2020comprehensive}.

\begin{figure}[htb]
    \centering
        \begin{tikzpicture}[shorten >=1pt, auto, node distance=1cm]
        \node[node_style] (v1) at (-2,0) {1};
        \node[node_style] (v2) at (-0.5,0.5) {2};
        \node[node_style] (v3) at (1.2,0) {3};
        \node[node_style] (v4) at (-1.5,2) {4};
        \node[node_style] (v5) at (0.8,2) {5};
        
        \draw[edge_style]  (v1) edge (v2);
        \draw[edge_style]  (v2) edge (v3);
        \draw[edge_style]  (v4) edge (v1);
        \draw[edge_style]  (v4) edge (v5);
        \draw[edge_style]  (v5) edge (v3);
        \draw[edge_style] (v4) edge (v2);
        
    \end{tikzpicture}
    \begin{tikzpicture}[shorten >=1pt, auto, node distance=1cm]
        \node[node_style_orange] (v1) at (-2,0) {1};
        \node[node_style_red] (v2) at (-0.5,0.5) {2};
        \node[node_style_orange] (v3) at (1.2,0) {3};
        \node[node_style_orange] (v4) at (-1.5,2) {4};
        \node[node_style] (v5) at (0.8,2) {5};
        \node[inner sep=0pt] (mail) at (0,0) {\includegraphics[width=.03\textwidth]{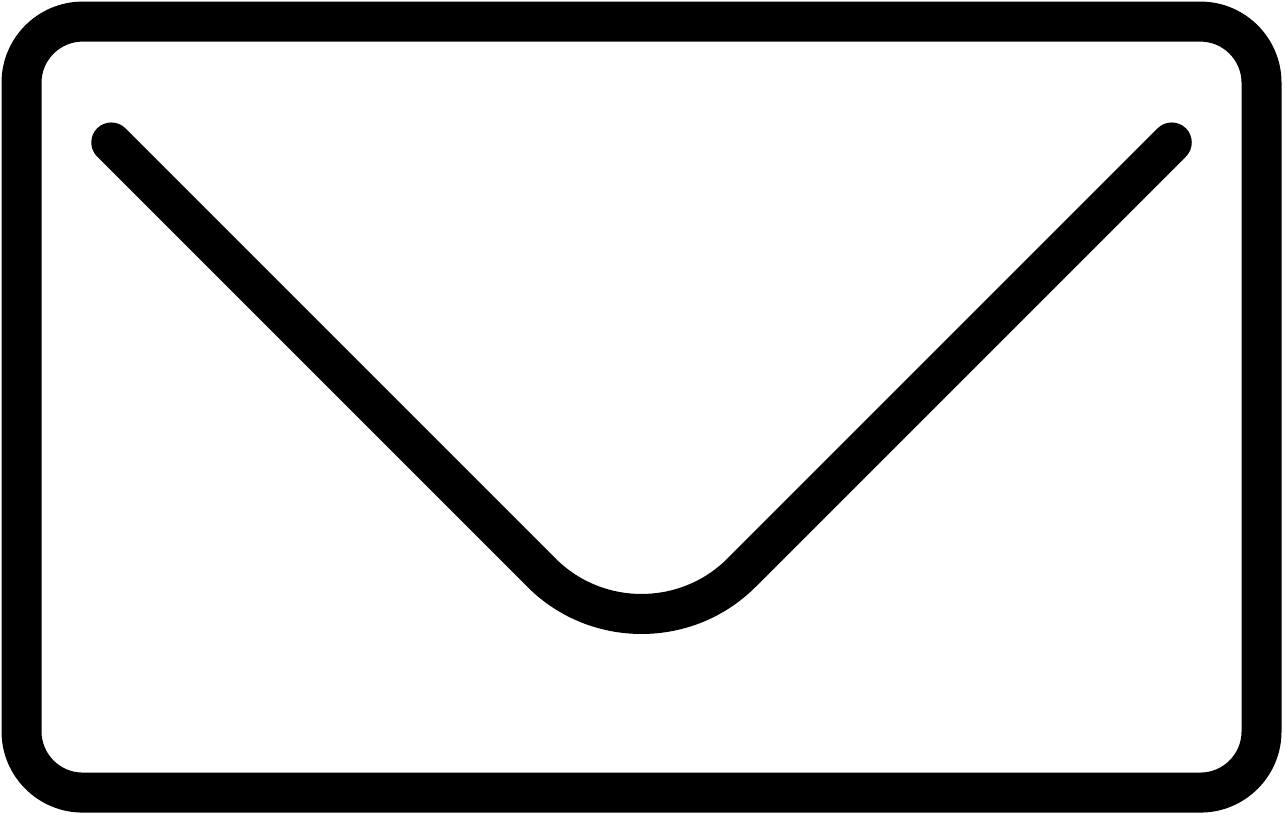}};
        \node[inner sep=0pt] (mail) at (-1.2,0) {\includegraphics[width=.03\textwidth]{mail.pdf}};
        \node[inner sep=0pt] (mail) at (-0.4,1.3) {\includegraphics[width=.03\textwidth]{mail.pdf}};
        
        \draw[edge_stylered, ->]  (v1) edge (v2);
        \draw[edge_stylered, ->]  (v3) edge (v2);
        \draw[edge_style]  (v4) edge (v1);
        \draw[edge_style]  (v4) edge (v5);
        \draw[edge_style]  (v5) edge (v3);
        
        \draw[edge_stylered, ->] (v2) to [out=380,in=430,looseness=5] (v2);
        \draw[edge_stylered, ->] (v4) edge (v2);
    
    \end{tikzpicture}
    \begin{tikzpicture}[shorten >=1pt, auto, node distance=1cm]
        \node[node_style_orange] (v1) at (-2,0) {1};
        \node[node_style_red] (v2) at (-0.5,0.5) {2};
        \node[node_style_orange] (v3) at (1.2,0) {3};
        \node[node_style_orange] (v4) at (-1.5,2) {4};
        \node[node_style] (v5) at (0.8,2) {5};
        \node[inner sep=0pt] (mail) at (0,0) {\includegraphics[width=.03\textwidth]{mail.pdf}};
        \node[inner sep=0pt] (mail) at (-1.2,0) {\includegraphics[width=.03\textwidth]{mail.pdf}};
        \node[inner sep=0pt] (mail) at (-0.4,1.3) {\includegraphics[width=.03\textwidth]{mail.pdf}};
        
        \draw[edge_stylered, ->]  (v2) edge (v1);
        \draw[edge_stylered, ->]  (v2) edge (v3);
        \draw[edge_style]  (v4) edge (v1);
        \draw[edge_style]  (v4) edge (v5);
        \draw[edge_style]  (v5) edge (v3);

        \draw[edge_stylered, ->] (v2) edge (v4);
    
    \end{tikzpicture}
    \caption{Example of the basic message passing procedure of a GNN/GCN.}
    \label{fig:propagation}
\end{figure}

Essentially, almost all the GNNs can be expressed as Message Passing Neural Networks \citep{gilmer2017neural}.
\begin{enumerate}
\item The message passing function defines how the convolution works;
\item a node update function determines the new node states after propagation;
\item a readout function determines what is done with this information (e.g., node classification or link prediction).
\end{enumerate}
Figure~\ref{fig:propagation} depicts a simple schematic overview of the node updating procedure, and its respective steps. First, node 2 will collect the node feature information from its neighbors. Then, it will update its state and also provide a message for its own neighbors, concluding the proposition of \citet{gilmer2017neural}.

The spectral-based GNNs exploit the adjacency or laplacian matrix and the degree matrix of a graph to perform the convolution in the Fourier domain, similar to GSP techniques. A graph signal is convoluted throughout the graph in the Fourier domain, and reversely transformed back to the graph domain. However, a severe limitation of spectral-based methods is the lack of Transferability, since the method is dependent on the specific graph it is trained on. Therefore, the graph neural network models also needs the entire graph to train on, which is more complex in larger graph settings.  

Spatial-based methods define the graph convolution on the spatial relations of a node, similar to the convolution step in a conventional CNN with image data. The graph convolution combines the representation of the central node's representation with its neighbors representations to derive the updated state of the central node. The spatial graph convolutional operation fundamentally propagates node information along edges. Below, we summarize the core mechanisms which are relevant for the methods applied in this paper.

In both types of convolutions, added to this propagation of information are optional node and edge features. 
These node features and edge features in a graph $G=(V,E)$ are the feature description $x_{i}$ for every node $i$ in the $V \times F$ matrix $X$, where $F$ is the number of input features.

However, to work with an adjacency matrix and to use node features and edge features, some adaptions have to be made to the classical way a neural network performs feature propagation. In normal neural networks, we propagate to the next layer by:

\begin{equation}
    H^{i+1}=\sigma \left(W^{i}H^{i}+b^{i}\right ),
\end{equation}

where $H^{i}$ is the feature representation of each node at layer $i+1$, $\sigma$ the activation function (e.g., Tanh or ReLU), $W^{i}$ the weights at layer $i$, $H^{i}$ the feature representation at layer $i$ and $b^{i}$ the bias at layer $i$.

\citet{kipf2016semi} formalized the propagation rule in a GCN as:

\begin{equation}
    H^{i+1} = \sigma\left( \hat{D}^{-\frac{1}{2}}\hat{A}\hat{D}^{-\frac{1}{2}}H^{(l)}W^{(l)}\right)
\end{equation}

where $W^{l}$ is the weight matrix, $\sigma$ the activation function (e.g., ReLU) and $\hat{A}$ the normalized adjacency matrix with the addition of the identity matrix $I$ and multiplying by the inverse degree matrix $\hat{D}$ of $\hat{A}$. These adaptions of the adjacency matrix and the degree matrix are necessary because of two reasons:
\begin{enumerate}
\item If we would multiply with the normal adjacency matrix $A$, then for every node, we would sum up all the neighboring nodes except the node itself. Adding the identity matrix $I$ to $A$ will ensure that the node features of the node in question will also be taken into account.

\item If the adjacency matrix $A$ would not be normalized, then nodes with a high degree will change the scale of the feature vectors. Once we use the symmetrically normalized adjacency matrix $D^{-\frac{1}{2}} A D^{-\frac{1}{2}}$, this problem is solved and the average of the neighboring nodes is used.
\end{enumerate}

The resulting representation is a vector-form that can be directly used for several tasks. For example, the features can be used to predict the labels of specific nodes in a graph. Another example is applying classification of the entire graph. In this paper, we use the information to improve the results of forecasting the strain at each node in the sensor network.

\section{Method}\label{sec:method} %
This section first provides an outline of our analysis framework. Then, the dataset and network modeling techniques are explained, before we describe the respective GSP and GNN methods.

\subsection{\emph{Overview: GSP Methodological Framework}}

\begin{figure}[htb]
 \centering
 \includegraphics[width=1\columnwidth]{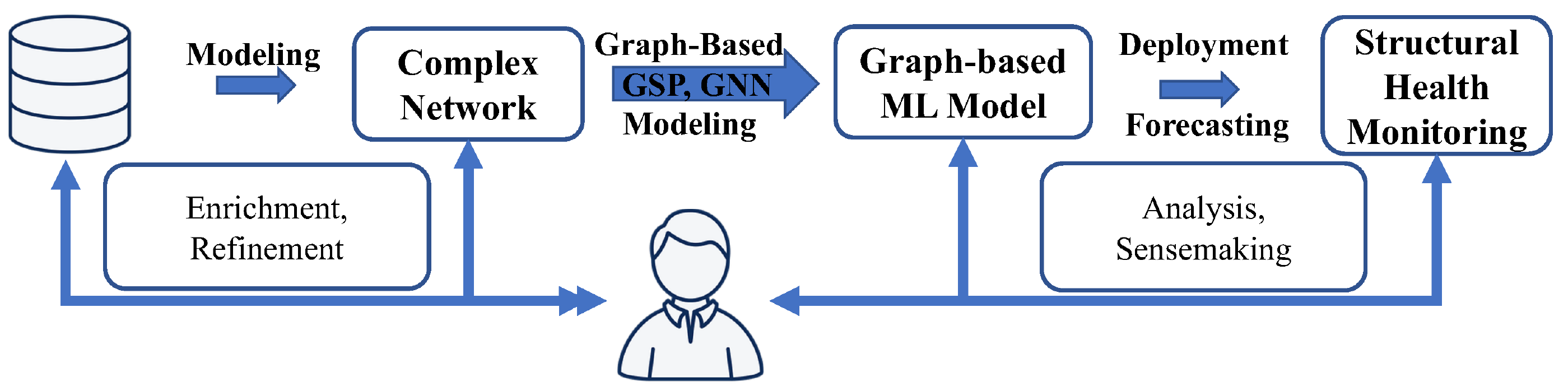}
 \caption{A brief description of the proposed Graph Signal Processing and Graph Neural Network framework for Structural Health Monitoring.} \label{fig:methodology}
\end{figure}

An overview on the proposed computational framework for modeling complex sensor network data using GSP/GNN in SHM is given in Figure~\ref{fig:methodology}; the figure depicts the overall processing and modeling pipeline of the framework. It is easy to see, that the presented framework provides an incremental and iterative workflow and methodology with a human-in-the-loop.
For the respective steps, of the framework, we proceed as follows:
\begin{enumerate}
    \item \emph{Input data:} We start with the data as input for the \emph{modeling step}.
    \item \emph{Modeling:} In the modeling step, we abstract the complex signal (\ie the sensor network data) into a \emph{complex network} representation.
    \item Already, at this step, the complex network model can be evaluated semi-automatically in order to add refinements and adaptations of the network, for example, when faulty sensor (data) is detected.
    \item \emph{Learning/Modeling:} Next, we apply GSP and GNN learning on our network model, in order to obtain a graph-based machine learning (ML) model.
    \item The resulting model can then be deployed for analysis and forecasting, in the context of SHM. Example applications include the identification of a minimal sensor subset, the detection of specific patterns, events, or mode shapes, as well as predicting specific diagnostic values - \eg strain etc. 
\end{enumerate}

Below, we will exemplify the application of this framework in more detail in our case study.

\subsection{Dataset}\label{sec:dataset}
The \emph{InfraWatch} project investigated the \emph{Hollandse Brug} (built in 1969), a large highway bridge in the Netherlands that connects the provinces of Noord-Holland and Flevoland.
After reports indicated that the bridge did not meet the quality and security requirements, sensors were placed at several locations on the bridge. This network of sensors includes 145 sensors, which contains 20 temperature, 41 vertical strain (Y-strain), 50 horizontal strain (X-strain) and 34 vibration sensors. Various data mining techniques have been applied to the dataset, including time series analysis \citep{vespier2012mdl} and modal analysis \citep{miao2013modal}.

The dataset that was made available to us includes 5 minutes of sensor data collected in high-resolution, approximately 30,000 observations in total. The original provided data was sampled at 100Hz. For smoothing the signal, we took the averaged values per 100ms. The data consists of several traffic events, where the 10 most significant are examined in this paper.

Our domain specialist suggested that the strain sensors were not measured on the same scale or at the same time. Since time synchronization is in general a challenging task when gathering simultaneous sensor data~\citep{mechitov2004high}, the clock times were matched by comparing the sensor reading peaks. Afterwards, the data was normalized by rescaling them using a standard z-score standardization method.

\begin{figure}[htb]
    \centering
    \begin{tikzpicture}[spy using outlines={circle,red,very thick,magnification=5,size=1.5cm, connect spies}]
    \node {\includegraphics[width=0.5\textwidth]{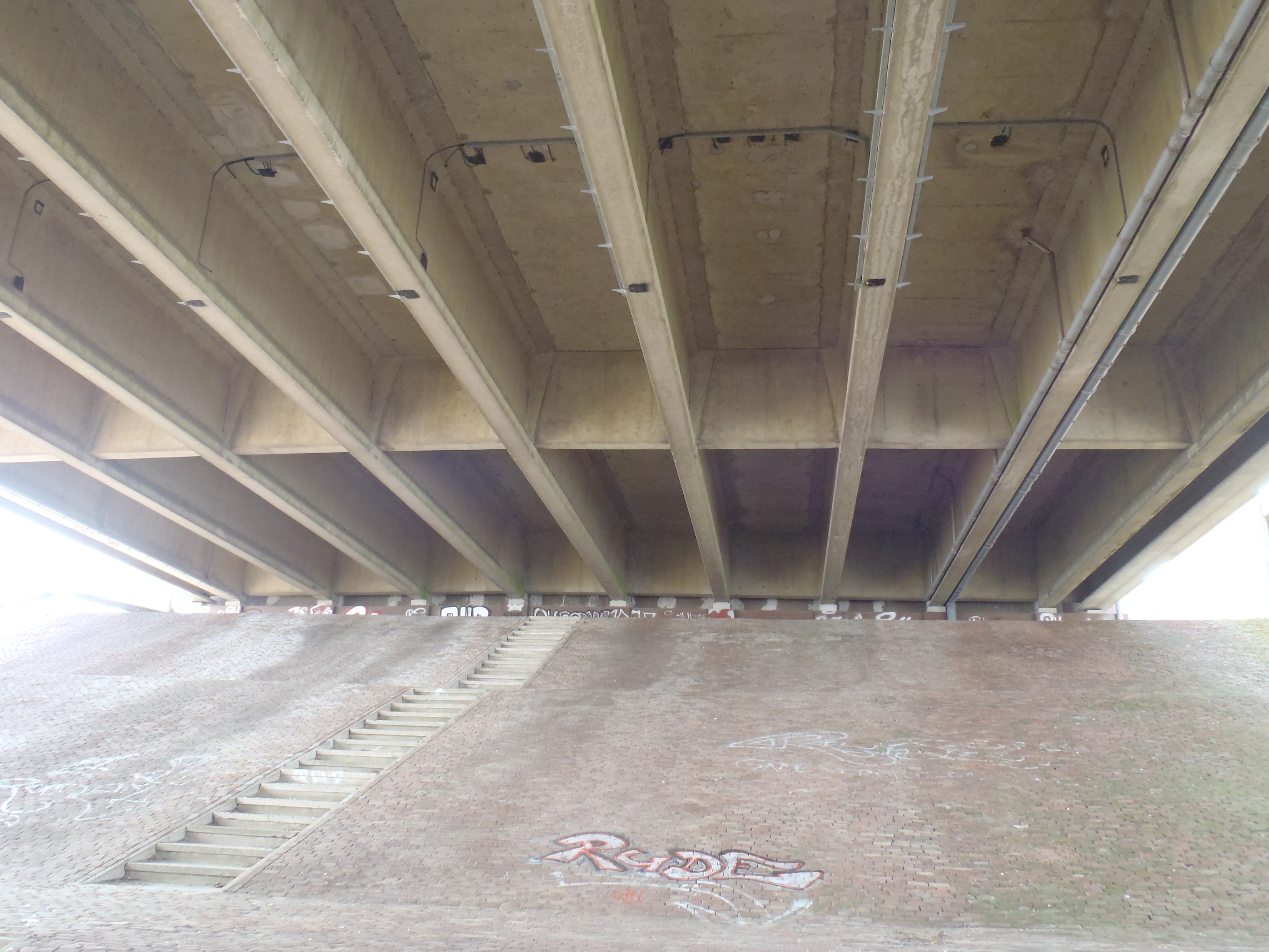}};
    \spy on (0,0.88) in node [left] at (2,1.25);
    \end{tikzpicture}
    \caption{Picture that highlights the placement of the sensors that are attached to the girders and bottom of the deck (from~\citep{miao2014structural}).}
    \label{fig:sensorphoto}
\end{figure}

\begin{figure}[htb]
    \centering
    \begin{tikzpicture}
    \node {\includegraphics[width=0.99\textwidth]{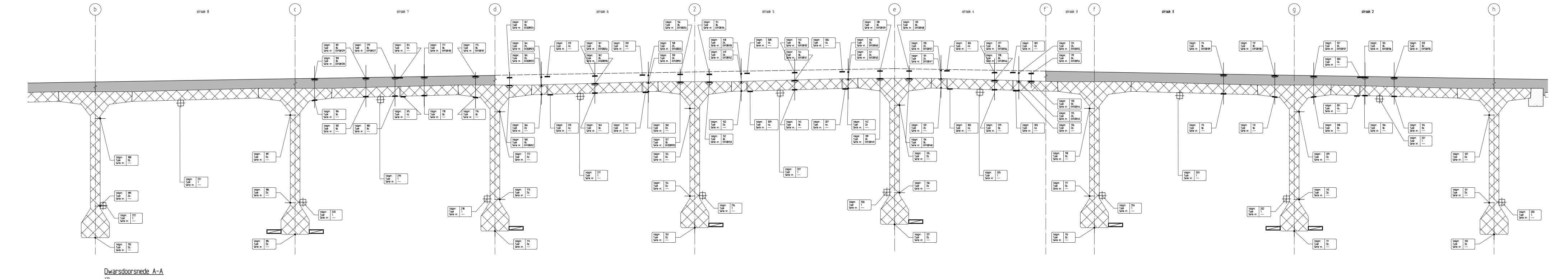}};
    \draw[color=red, very thick](0.82,-0.65) circle (0.2);
    \end{tikzpicture}
    \caption{Blueprint of the sensors locations. The red circle shows the location of the same sensor highlighted in Figure \ref{fig:sensorphoto} on the bridge.}
    \label{fig:bridgephoto}
\end{figure}

The sensors were mounted at three different cross-sections within a single span, \cf Figures~\ref{fig:sensorphoto}-\ref{fig:bridgephoto} (see~\citep{miao2014structural} for more visual information). 
As a result, in order to make the network links relevant, the 31 sensors in the middle and right cross-sections were removed. 
In addition, four sensors were discovered to be unreliable, reducing the total number of sensors from 145 to 110.

\subsection{Network Creation}
The blueprint of the bridge provides the geographical locations of the sensors to create the network. The choice for each edge $(i,j)$ is a bit more difficult, but also a crucial step \citep{mateos2019connecting}. A possible direction could be using geographical distance, but that would not grasp the functional relationship between the sensors, since the girders on the bridge should catch most of the strain. As a result, although the sensors at the top of the bridge are geographically similar to the other sensors, they should behave in the exact opposite manner as the strain sensors at the girders. Therefore, the edges were determined by either (1) the correlation score or (2) the Dynamic Time Warp (DTW) distance between the sensor readings. Lastly, only the top three edges with the highest weight were added to the network (excluding the vibrations sensors, which had few edges in the first place).

The DTW distance can by calculated by first dividing time series 1 and time series 2 into equal points. Afterwards, the euclidean distance is calculated between each point in the first time series and each point in the second, where the minimum distance is stored. This procedure is repeated for every point in the first time series until all data points are evaluated. The sum of all the minimum distances is then the measure of similarity between the two series.

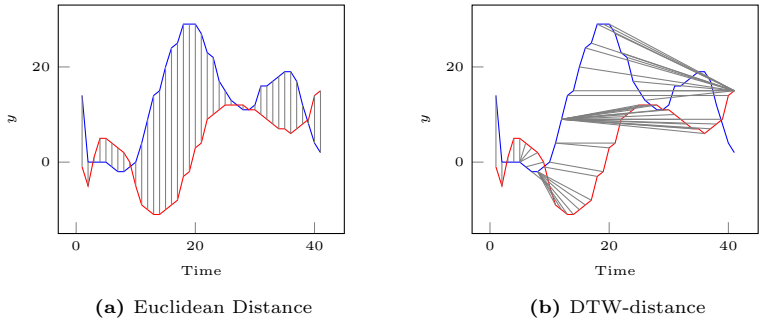
\begin{figure}[htb]
\centering 
  \begin{subfigure}[b]{0.45\textwidth}
    \begin{tikzpicture}
        \begin{axis}[
            xlabel=Time,
            width=\textwidth,]
            \addplot[blue, mark=none] table[x=t, y=q] {timeseries.dat};
            \addplot[red, mark=none] table[x=t, y=r] {timeseries.dat};
            \addplot[gray, quiver={v=\thisrow{v}}] table[x=t, y=q] {timeseries.dat};
        \end{axis}
    \end{tikzpicture}%
    \caption{Euclidean Distance} \label{fig:M1}  
  \end{subfigure}
\begin{subfigure}[b]{0.45\textwidth}
  \begin{tikzpicture}
        \begin{axis}[
            xlabel=Time,
            width=\textwidth,]
            \addplot[blue, mark=none] table[x=t, y=q] {timeseries.dat};
            \addplot[red, mark=none] table[x=t, y=r] {timeseries.dat};
            \addplot[gray, quiver={u=\thisrow{u}, v=\thisrow{v}}] table[x=t, y=r] {path.dat};
        \end{axis}
    \end{tikzpicture}
\caption{DTW-distance} \label{fig:M2}  
\end{subfigure}
\caption{Difference between the euclidean distance and the DTW-distance. The DTW method allows more flexibility in measuring time series that do not sync perfectly in time.}
\end{figure}  

It is interesting to investigate which of the two techniques is most suited for each method used in this paper. To start, the main difference between both techniques is that DTW assumes that each time series is on the same scale, while correlation is scale-invariant. On the other hand, correlation is a more global-based measure, which means that information signalling direction (one time series causing effect in the other) is not available. Therefore, each technique has advantages in different situations over the other.

To conclude, several networks are created. First, the X-strain network with 42 sensors and 126 edges. Second, the Y-strain network with 37 sensors and 111 edges. Third, a combination of X and Y sensors with 79 nodes and 237 edges. Lastly, the vibration sensors form an especially small graph with 15 nodes and 26 edges. Each network had their own contribution to the analysis of this paper. For example, the strain sensors were used to conduct the sampling, mode shape identification and forecasting. The vibration sensors mainly assisted with identification of the mode shapes.

\subsection{Node Subset Selection -- Sensor Subset Sampling}
\begin{figure}[b]
    \centering
    \includegraphics[width=\textwidth]{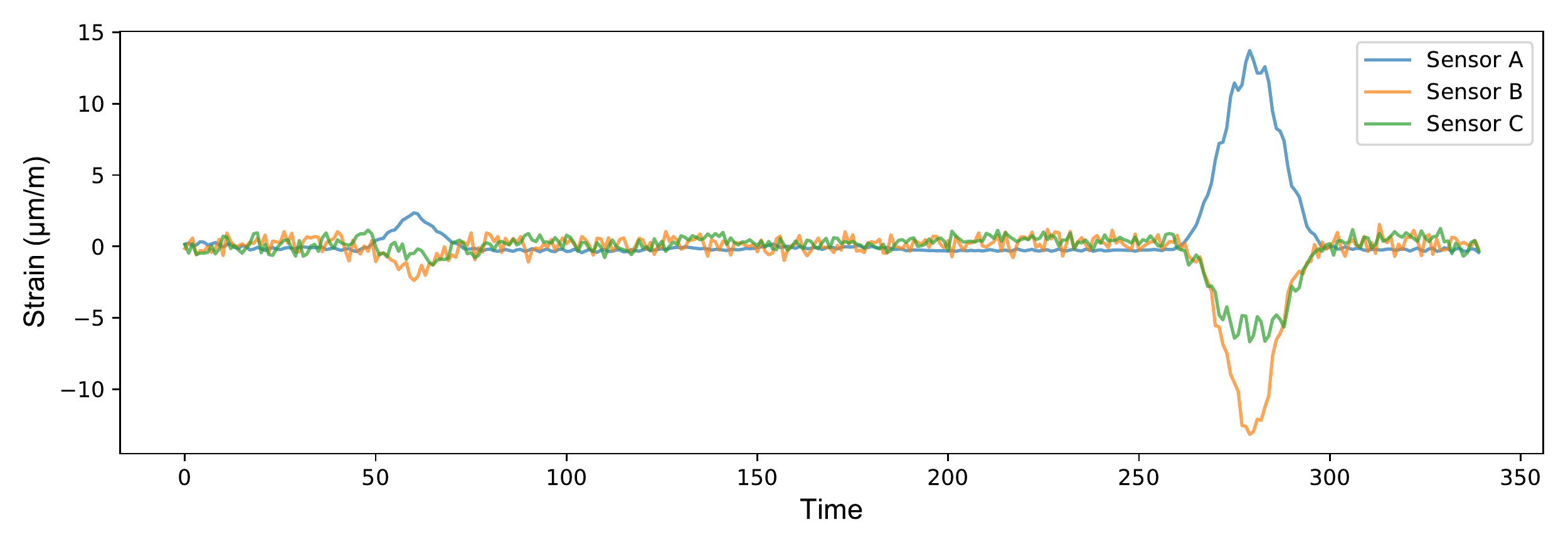}
    \caption{Strain values ($\mu\textrm{m/m}$) of three sensors attached to the bridge at different locations. Sensor B and C show how the strain behaves differently, influenced by their position on the bridge. Sensor A is placed at the right-most girder on the bridge; it captures most of the strain, whereas Sensor B and C are placed on the deck of the bridge where the strain is countered.} 
    \label{fig:allstrain}
\end{figure}

One of the core tasks in GSP is to ``reconstruct'' the signal of sensors, \ie deducing those given a specific sample. As an example, consider a case in which cost or bandwidth limits restrict the number of applied sensor nodes.
In this work, we apply sampling by calculating the optimal subset of sensors that are able to reconstruct the original signal at specific time points. These specific moments in time refer to the traffic events that happen on the bridge, since calculating the error during no traffic would inflate our reconstruction error. Figure \ref{fig:allstrain} motivates this decision, where the time points outside of the peaks show little deviation at all. Incorporating these time points in the evaluation will highly influence and skew the results of each algorithm. 

Since brute-force searching for the most optimal solution is not feasible with a large number of $N$ nodes, the following strategies are investigated: random search and top-down or bottom-up hill-climbing. Both last strategies are well known greedy-search strategies~\citep{krause2008near,aggarwal2017sensor,anis2016efficient,puy2018random}. The random search strategy serves as a baseline and creates a random set of sensors that are sampled. On the contrary, the hill climbing algorithms either perform subset selection in a bottom-up or top-down manner. Bottom-up hill climbing (i.e., Forward Selection) starts with zero selected sensors and progressively selects the most informative sensors that decrease the error the most. Top-down hill climbing (i.e., Backward Elimination) consist of starting with all sensors and eliminating the least informative sensors one-by-one. Both hill climbing techniques incorporate a random element by choosing from the top-3 either best or worst performing sensors, which helps to prevent the hill climbers to reach local maxima and minima. Each algorithm ran a total of 500 iterations to find the best solution and was terminated when 25\% of the sensors were selected. It could then occur that the same result is found multiple times, so only the unique solutions were stored.

To estimate the original signal from this subset of sensors, Tikhonov Minimization is applied in each iteration of the sampling procedure~\citep{shuman2013emerging,pygsp}.
The function solves for the unknown vector $x$:

\begin{equation}\label{eq:tikhonov1}
    \textrm{arg}\min_x ||Mx-y||\frac{2}{2}+\tau x^{T}Lx,
\end{equation}

\noindent if $\tau > 0$ and 

\begin{equation}\label{eq:tikhonov2}
    \textrm{arg}\min_x x^{T}Lx : y = Mx,
\end{equation}

\noindent otherwise, with the graph signal $y$, the masking vector $M$ that resembles a binary vector of which nodes are sampled (1 = sampled, 0 = not sampled), the laplacian matrix $L$ and the regularization parameter $\tau$. 
Several values for the regularization parameter in the Tikhonov Minimization were tried, of which the default value of $\tau = 0$ showed the optimum results.

Finally, each algorithm was also applied on data obtained by applying a low-pass filter $g(x)$ on the graph frequencies $x$ of the signals: $$g(x) = \frac{1}{1 + 0.5 \cdot x}\,.$$ Essentially, a graph filter performs a transformation as a function over the graph frequencies (in our example $g(x)$); it alters their contents by a point-wise multiplication in the graph Fourier domain~\citep{isufi2019graph}. After the filter has been applied, the Inverse Graph Fourier Transformation of the Fourier domain signal reverts the signal back to the time domain for evaluation.

\subsection{Strain Forecasting Method}
In order to forecast the strain values in the sensor data, we applied the T-GCN utilizing the implementation contained in the Stellargraph Package~\citep{StellarGraph}. The T-GCN is a Spatio-Temporal Graph Convolutional Network that combines graph layers with Long Short-Term Memory layers \citep{zhao2019t}. The spatial aspect of the data lies in the exact locations of the sensors, whereas the temporal aspect lies in the fact that different loads over time produce different stress on the bridge. For example, there could be a daily pattern in the direction of the traffic on the bridge.

The strain data of each sensor type was cut into 2914 sequences of length 10 (10 x 100ms), where the task was to predict the strain value at the 12th timestep in the future (\ie the most difficult setting in the original T-GCN paper by~\cite{zhao2019t}). In other words, we estimate the values 1.2 seconds later based on the preceding 1 second of strain observations. In our experimentation, we used an 80/20 split for training and testing. The model used the $N \times N$ adjacency matrix and the $N \times T$ feature matrix $X$, which describes the strain over $T$ timesteps for $N$ sensors. In this way, we regard the strain values $X_{t} \in R^{N \times T}$ the strain measured at each sensor on the bridge at time $i$. We can thus consider our problem definition as learning the function $f$ on the network topology of the sensor graph $G$ and the feature matrix $X$ to calculate the strain at timestep $t$.

For setting up the T-GCN, two GCN layers were used with each 8 filters. These were attached to two Long Short-Term Memory (LSTM) layers with each 50 filters. LSTM layers are special recurrent layers where the top horizontal line $C_{t}$ is the memory state, enabling the LSTM to remember information from the past. It also contains gates that allow or block information in the network from passing by, and these gates consist of Sigmoid functions and multiplication operations. For example, the first sigma gate in Figure \ref{fig:lstm} functions as a \emph{Forget Gate}, blocking or allowing information to flow through. The second sigma functions as an input gate and the third sigma functions as the output gate. 
Lastly, the dense output layers consist of the $N$ sensors in the graph with Tanh as the activation function, since the strain values can be negative and fall between [-1,1]. To conclude, Table \ref{tab:ModelOverview} shows an overview of the T-GCN model. To calculate the quality of the predictions, the Root Mean Squared Error (RMSE) of each segment is taken and compared to a benchmark taking the most recently observed value. Such a benchmark is tough to beat, since it is not expected that the strain signal will differ significantly in a short period of time.

\newcommand{\empt}[2]{$#1^{\langle #2 \rangle}$}
\usetikzlibrary{positioning, fit, arrows.meta, shapes}

\begin{figure}[ht]
    \centering
    \resizebox{0.5\textwidth}{!}{%
    \begin{tikzpicture}[
    font=\sf \scriptsize,
    >=LaTeX,
    cell/.style={
        rectangle, 
        rounded corners=5mm, 
        draw,
        very thick,
        },
    operator/.style={
        circle,
        draw,
        inner sep=-0.5pt,
        minimum height =.2cm,
        },
    function/.style={
        ellipse,
        draw,
        inner sep=1pt
        },
    ct/.style={
        circle,
        draw,
        line width = .75pt,
        minimum width=1cm,
        inner sep=1pt,
        },
    gt/.style={
        rectangle,
        draw,
        minimum width=4mm,
        minimum height=3mm,
        inner sep=1pt
        },
    mylabel/.style={
        font=\scriptsize\sffamily
        },
    ArrowC1/.style={
        rounded corners=.25cm,
        thick,
        },
    ArrowC2/.style={
        rounded corners=.5cm,
        thick,
        },
    ]

    \node [cell, minimum height =4cm, minimum width=6cm, fill=green!10] at (0,0){} ;

    \node [gt, fill=yellow!30] (ibox1) at (-2,-0.75) {$\sigma$};
    \node [gt, fill=yellow!30] (ibox2) at (-1.5,-0.75) {$\sigma$};
    \node [gt, minimum width=1cm, fill=yellow!30] (ibox3) at (-0.5,-0.75) {Tanh};
    \node [gt, fill=yellow!30] (ibox4) at (0.5,-0.75) {$\sigma$};

    \node [operator, fill=purple!30] (mux1) at (-2,1.5) {$\times$};
    \node [operator, fill=purple!30] (add1) at (-0.5,1.5) {+};
    \node [operator, fill=purple!30] (mux2) at (-0.5,0) {$\times$};
    \node [operator, fill=purple!30] (mux3) at (1.5,0) {$\times$};
    \node [function, fill=purple!30] (func1) at (1.5,0.75) {Tanh};

    \node[ct] (c) at (-4,1.5) {$C_{t-1}$};
    \node[ct] (h) at (-4,-1.5) {$H_{t-1}$};
    \node[ct] (x) at (-2.5,-3) {$X_{t}$};

    \node[ct] (c2) at (4,1.5) {$C_{t}$};
    \node[ct] (h2) at (4,-1.5) {$H_{t2}$};
    \node[ct] (x2) at (2.5,3) {$H_{t1}$};

    \draw [ArrowC1] (c) -- (mux1) -- (add1) -- (c2);

    \draw [ArrowC2] (h) -| (ibox4);
    \draw [ArrowC1] (h -| ibox1)++(-0.5,0) -| (ibox1); 
    \draw [ArrowC1] (h -| ibox2)++(-0.5,0) -| (ibox2);
    \draw [ArrowC1] (h -| ibox3)++(-0.5,0) -| (ibox3);
    \draw [ArrowC1] (x) -- (x |- h)-| (ibox3);

    \draw [->, ArrowC2] (ibox1) -- (mux1);
    \draw [->, ArrowC2] (ibox2) |- (mux2);
    \draw [->, ArrowC2] (ibox3) -- (mux2);
    \draw [->, ArrowC2] (ibox4) |- (mux3);
    \draw [->, ArrowC2] (mux2) -- (add1);
    \draw [->, ArrowC1] (add1 -| func1)++(-0.5,0) -| (func1);
    \draw [->, ArrowC2] (func1) -- (mux3);

    \draw [-, ArrowC2] (mux3) |- (h2);
    \draw (c2 -| x2) ++(0,-0.1) coordinate (i1);
    \draw [-, ArrowC2] (h2 -| x2)++(-0.5,0) -| (i1);
    \draw [-, ArrowC2] (i1)++(0,0.2) -- (x2);

\end{tikzpicture}
}%
    \caption{Schematic overview of a single LSTM cell.}
    \label{fig:lstm}
\end{figure}
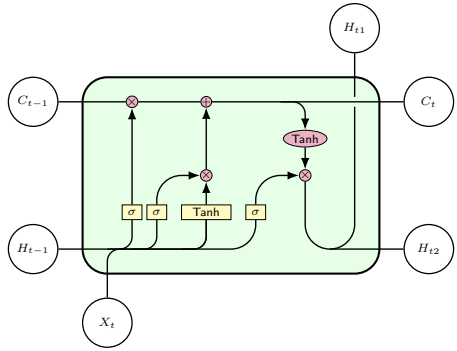

\begin{table}
\centering
\caption{Schematic overview of the architecture of the T-GCN on the X-strain data with 42 nodes.}
\label{tab:ModelOverview}
\begin{tabular}{@{}lcccc@{}}
\toprule
Layer & Activation & Filters & Shape & Parameters \\ \midrule
Input & - & - & 42 x 10 & 0 \\
GCN 1 & ReLU & 8 & 42 x 8 & 1886 \\
GCN 2 & ReLU & 8 & 42 x 8 &  1870 \\
\cmidrule{1-5}
Reshape 1 & - & - &  42 x 8 x 1 &   0 \\
Permute & - & - & 8 x 42 x 1 & 0 \\
Reshape 2 & - & - & 8 x 42 & 0 \\
\cmidrule{1-5}
LSTM 1 & Tanh & 50 & 8 x 50 &  18600 \\
LSTM 2 & Tanh & 50 & 50 &  20200 \\
Dropout & - & - &  50 &  0 \\
\cmidrule{1-5}
Dense & Tanh & 42 & 42 &  2142 \\ \bottomrule
\end{tabular}
\end{table}

\section{Results and Discussion}\label{sec:results}
Below, we present and analyze the results of the sensor sampling from which the total signal can be reconstructed. 
Then, mode shape detection and forecasting applications will be discussed.

\subsection{Sampling: Selecting a Minimal Subset of Sensors}
To select the minimal subset of sensors, we used three algorithms: (1) random selection and (2) bottom-up or (3) top-down hill-climbers and assessed their performance based on the Root Mean Squared Error (RMSE) scores. We chose RMSE as a metric since it (1) penalizes errors more than other metrics and (2) measures in the same unit as the variable of interest. Each algorithm was also tested on either the correlation-based graph or the DTW-based graph during the most noticeable traffic events (see Table~\ref{tab:newtesttable}). 

In terms of RMSE, the top-down algorithm continuously beats the random (+28.39\%) and bottom-up (+11.41\%) algorithms.
In addition, the random algorithm was tested for 50.000 iterations in a separate experiment (100x the original setting). Even after so many runs, the random algorithm did not outperform both hill-climbers when individual events were considered.
In this way, a \emph{single} top-down iteration outperforms a large number of random iterations (for any reasonable amount of iterations). 

Considering the type of sensors, our domain specialist indicated that the bridge can travel more freely in the Y-direction. Therefore, modeling the Y-strain could be more difficult than the X-strain.
Our results support this intuition, showing that the algorithms work well with X-strain sensors but fail with Y-strain sensors.

The DTW-based graph outperforms (+10.5\%) the correlation-based graph in terms of overall subset recovery performance. In addition, the DTW-graph also shows a reduction in the standard deviation in the RMSE results of the Y-strain and Combined sensors. However, it is remarkable to see that while the correlation-based graph version scores best in the X-strain condition, such a pattern is not visible in the DTW-graph. In the DTW-graph, combining the X and Y-strain sensors yields the best results.

\begin{table}[ht]
\centering 
\caption{Mean and standard deviation of RMSE scores in the DTW-graph and correlation-graph for all traffic occurrences for every algorithm. The columns non-filtered and filtered show whether or not graph signal filtering was used.}
\label{tab:newtesttable}
\begin{tabular}{@{}lcccccccc@{}}
\toprule
Algorithm & \multicolumn{7}{c}{Sensor Type} \\ \cmidrule(l){2-8}
 & \multicolumn{3}{c}{Non-Filtered} &  & \multicolumn{3}{c}{Filtered} \\ \cmidrule(lr){2-4} \cmidrule(l){6-8}
 & X-strain & Y-strain & Combined &  & X-strain & Y-strain & Combined \\ \midrule
\textbf{Correlation} &      &      &      & &      &      &       \\ 
Random    & 0.80  & 1.36   & 1.12  &  & 0.45  & 0.86  & 0.68 \\
          & (\emph{.32}) & (\emph{.95})  & (\emph{.76}) &  & (\emph{.29}) & (\emph{.62}) & (\emph{.46}) \\ \noalign{\vskip 2mm}  
Top-Down  & 0.60  & 1.06   & 0.74  &  & 0.31  & 0.66  & 0.38 \\
          & (\emph{.24}) & (\emph{.75})  & (\emph{.52}) &  & (\emph{.19}) & (\emph{.51}) & (\emph{.29}) \\ \noalign{\vskip 2mm}
Bottom-Up & 0.68  & 1.08   & 0.88  &  & 0.34  & 0.71  & 0.46 \\
          & (\emph{.30}) & (\emph{.80})  & (\emph{.68}) &  & (\emph{.21}) & (\emph{.53}) & (\emph{.35}) \\ 
\noalign{\vskip 4mm}
\textbf{DTW} &      &      &      & &      &      &       \\ 
Random    & 1.07  & 0.98   & 1.09  &  & 0.47  & 0.50  & 0.48 \\
          & (\emph{.66}) & (\emph{.34})  & (\emph{.41}) &  & (\emph{.31}) & (\emph{.22}) & (\emph{.22}) \\ \noalign{\vskip 2mm}
Top-Down  & 0.76  & 0.76   & 0.64  &  & 0.31  & 0.37  & 0.29 \\
          & (\emph{.38}) & (\emph{.24})  & (\emph{.27}) &  & (\emph{.20}) & (\emph{.16}) & (\emph{.16}) \\ \noalign{\vskip 2mm}
Bottom-Up & 0.99  & 0.90   & 0.83  &  & 0.42  & 0.42  & 0.35 \\
          & (\emph{.65}) & (\emph{.29})  & (\emph{.37}) &  & (\emph{.32}) & (\emph{.16}) & (\emph{.18}) \\ \bottomrule
\end{tabular}
\end{table}

\begin{figure}[htb]
    \centering
    \begin{minipage}[t]{0.48\textwidth}
        \centering
        \includegraphics[width=1\textwidth]{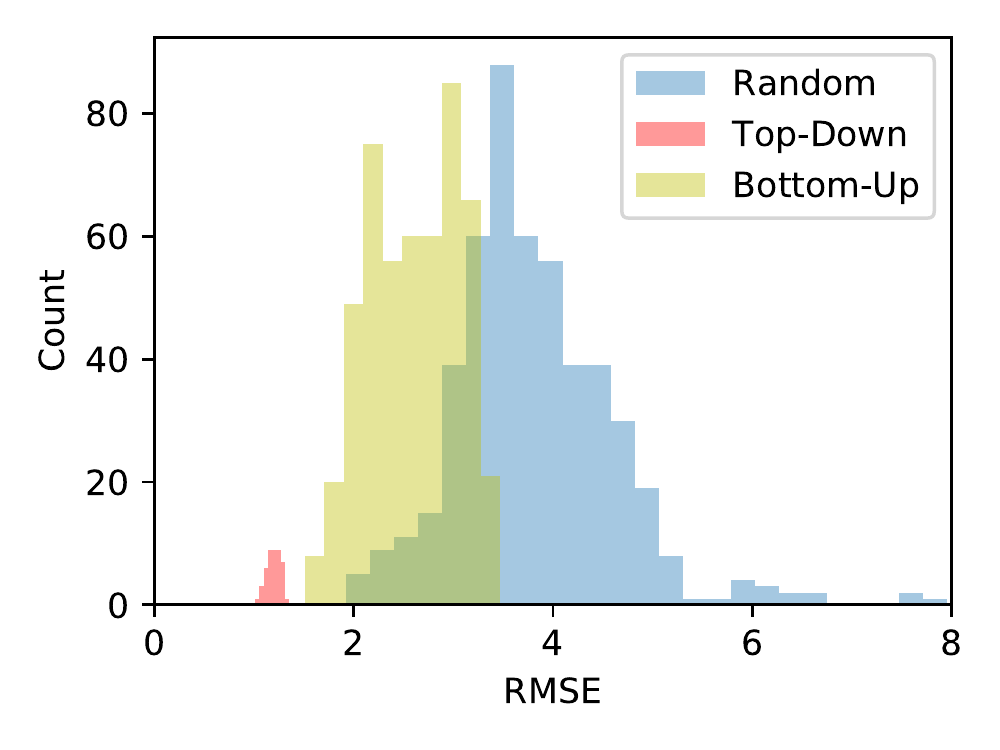} 
        \caption{RMSE scores of each algorithm on X-strain sensors during traffic event 1. in the correlation graph}
        \label{fig:algorithmsoverview}
    \end{minipage}\hfill
    \begin{minipage}[t]{0.48\textwidth}
        \centering
        \includegraphics[width=1\textwidth]{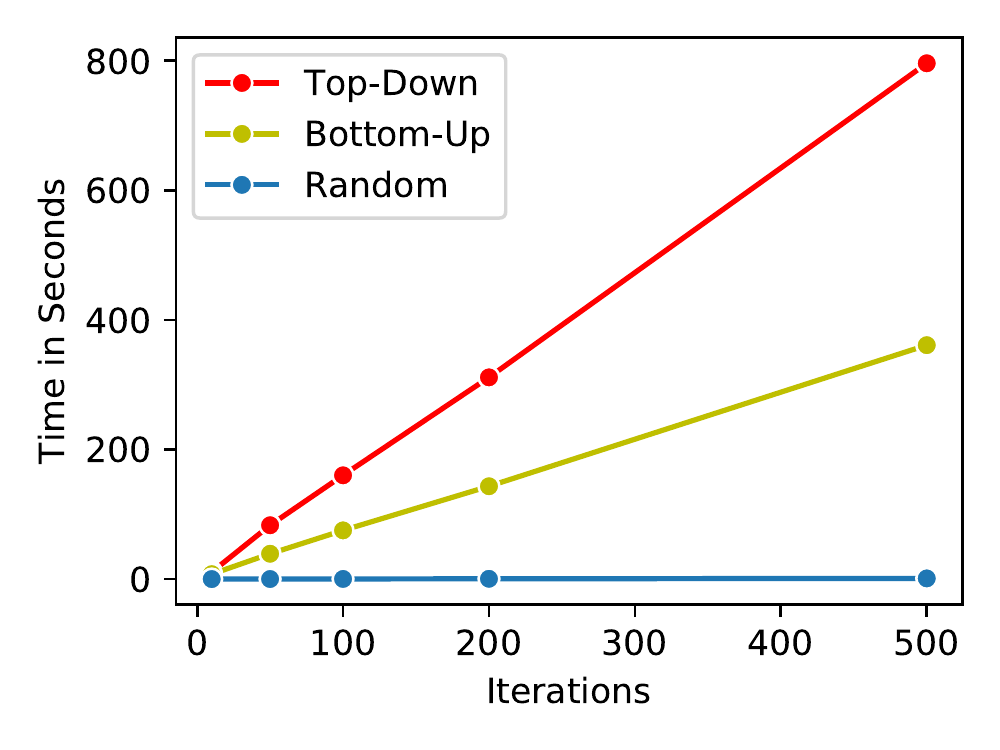} 
        \caption{Runtimes of each algorithm on X-strain sensors in the correlation graph.}
        \label{fig:convergence}
    \end{minipage}
\end{figure}

Considering the individual performance of each subset sampling algorithm, Figures~\ref{fig:algorithmsoverview}/\ref{fig:convergence} show the performance on the X-strain sensors of a specific traffic event in the correlation-based graph.
Set side by side, the top-down algorithm shows the best general performance (M = 1.20, {\it{SD}} = .06) and shares no overlap with the iterations of the bottom-up algorithm. The bottom-up algorithm, however, already improved from the random algorithm (M = 2.60, {\it{SD}} = .45), which performed worst.

The top-down algorithm also has a substantially lower standard deviation, indicating that it performs more consistently when multiple iterations are carried out. The fundamental procedures of the hill-climbers will shed light on such results. 
The bottom-up algorithm will calculate more unique iterations since the algorithm selects sensors instead of dropping them.
It calculates the 25\% selected sensors, while the top-down algorithm calculates the 75\% sensors \emph{not} selected. 
Therefore, weak sensors will almost always be removed in the top-down algorithm whereas such sensors could potentially still remain in the bottom-up algorithm \emph{longer}.
Therefore, running a few top-down trials to determine the ideal subset seems most optimal.

\begin{figure}[htb]
    \centering
    \includegraphics[width=1\textwidth]{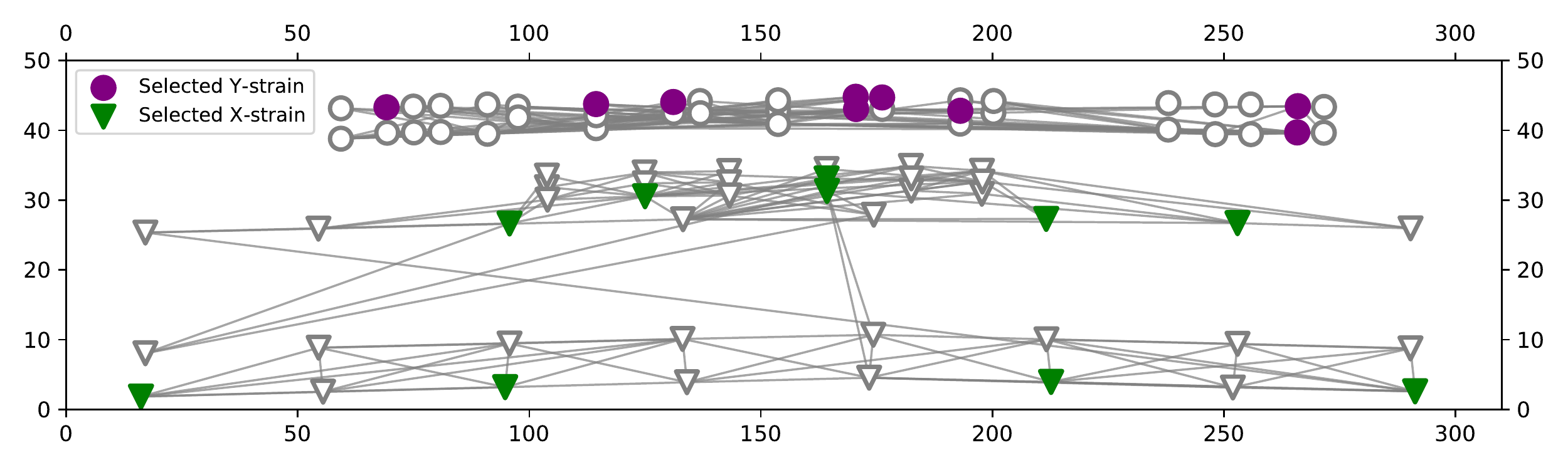}
    \caption{For both the X-strain and Y-strain sensors, the colored nodes are the sensors selected by the top-down algorithm.}
    \label{fig:coverageplot}
\end{figure}

Figure \ref{fig:coverageplot} shows a nearly symmetrical set of sensors chosen by the top-down algorithm.
Such a pattern is specifically noticeable in the X-strain sensors.
These results point to potential over-engineering in the number of sensors used on the bridge.
Furthermore, the second-lowest row of X-strain sensors (shown in Figure \ref{fig:coverageplot}) was not sampled at all.
This suggests that the sensors mounted in the center of the girders are not very useful, and could potentially be left out when designing future Condition Health Monitoring projects of bridges. Perhaps engineers could use the insights from such subset sampling techniques to determine the locations of the sensors in a data-driven way in future SHM projects.

\subsection{Network Representation Example: Girders and Deck}
\begin{figure}[ht]
    \centering
    \begin{minipage}[t]{0.48\textwidth}
        \centering
        \includegraphics[width=1\textwidth]{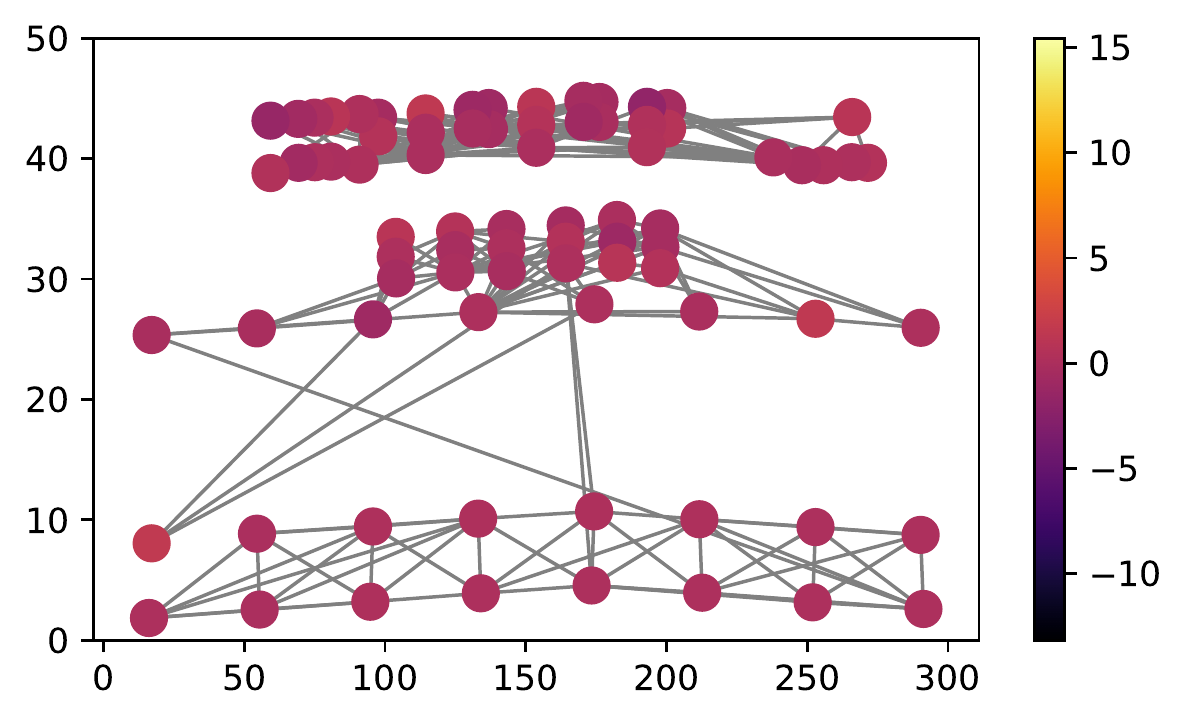} 
        \caption{The sensors network during normal conditions without traffic on the bridge.}
        \label{fig:xandy6}
    \end{minipage}\hfill
    \begin{minipage}[t]{0.48\textwidth}
        \centering
        \includegraphics[width=1\textwidth]{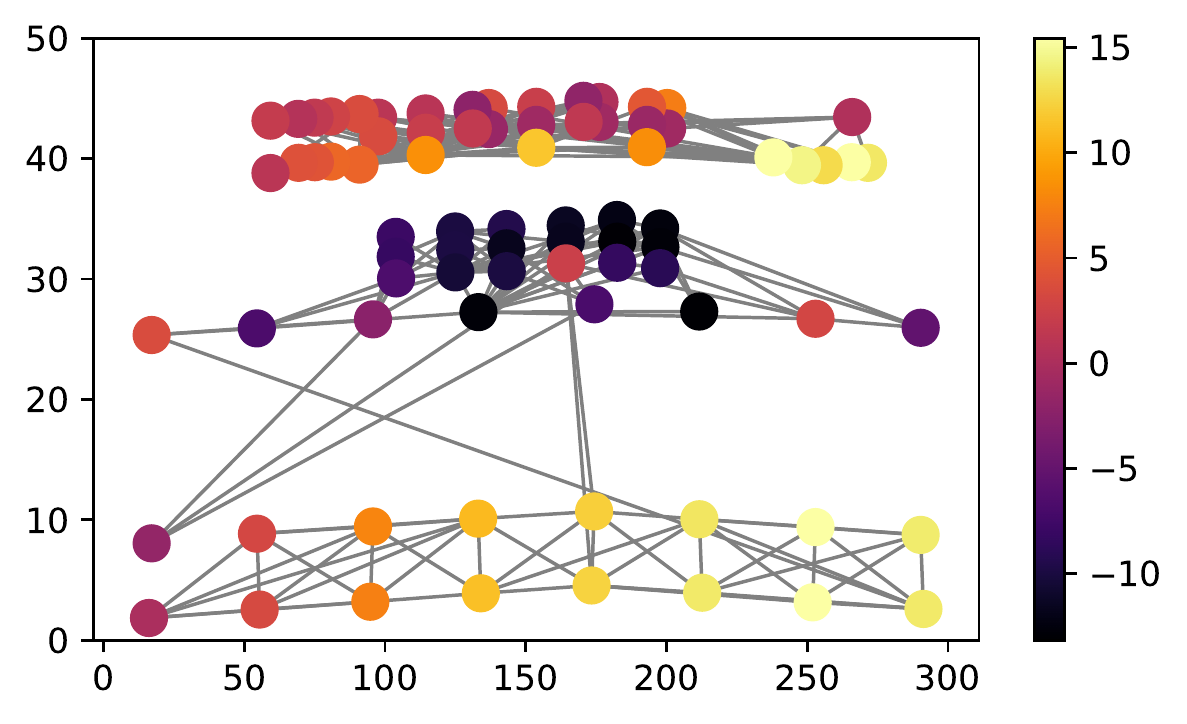} 
        \caption{The sensor network during a traffic event on the right side of the bridge. }
        \label{fig:xandy279}
    \end{minipage}
\end{figure}

For the network representation, we depict the sensors regarding X-strain and Y-strain in Figure \ref{fig:xandy6}; \ref{fig:xandy279} shows the respective sensors in one visualization. 
When examining Figure~\ref{fig:xandy6}, most of the connections in the network are between the strain sensors placed at either the top or the bottom of the bridge. 
Such a behavior is expected, which confirms our modeling choices: The bottom part of the bridge contains girders that carry most of the weight, which should all behave very differently from the sensors placed on top of the bridge deck. 
Figure~\ref{fig:xandy279} depicts traffic event 1, in which pressure can be seen on the bottom right side of the bridge, showing that one or more vehicle(s) passed by. 
The figure also shows a decline in strain located at the bridge deck, indicating that the girders are doing their function properly, according to our domain specialist.
Engineers could track the signals over time and determine how the pressure and vibrations are transmitted through the bridge. Figure~\ref{fig:xandy279} also shows a misbehaving sensor placed in the middle of the graph, of which engineers could assess whether this behavior is expected (placed on a special spot on the bridge) or not.

\begin{figure}[ht]
    \centering
    \includegraphics[width=0.55\textwidth]{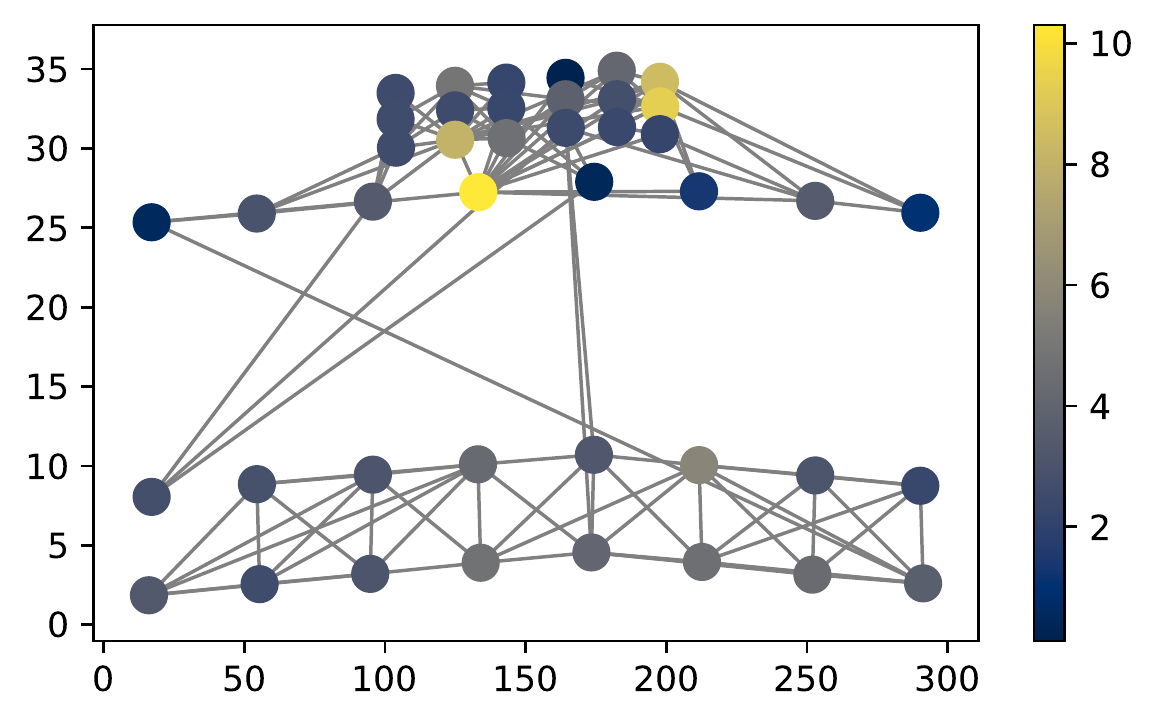}
    \caption{Total variation during a traffic event on the X-strain sensor graph. The yellow colored nodes are the nodes whose sensor readings differ most from their neighbors.}
    \label{fig:totalvariation}
\end{figure}

Figure~\ref{fig:totalvariation} helps us investigate the behavior of the sensor network in more detail. 
The total variation of each node during a traffic event relative to the sensors to which it is connected is plotted as a signal on the X-strain network.
The sensors placed on the girders indicate that the strain is equally distributed across the constructed girders. 
However, there is a lot more variation in the strain on the deck of the bridge. The most yellow-colored sensors are the sensors that highlight this behavior. The total variation could be used in a global manner as a measure for signal smoothness, whereas the localized version could highlight inaccurate sensor readings or sensor replacement. 

\subsection{Identification of Mode Shapes}
The Finite Element Method (FEM) is a computational technique for solving partial differential equations. It is commonly used to categorize the frequencies of signals into a combination of different modes in order to distinguish mode shapes. In general, FEM can be applied on any physical phenomenon, \eg heat flow, fluid behavior and wave propagation. FEM tries to solve a problem by partitioning a system into a set of smaller parts. These parts are the so-called ``finite elements'', which basically act as a representation of the entire object. Each element contains a simple equation that when combined models the global problem. 

\begin{figure*}[ht]
    \centering
    \begin{subfigure}[b]{0.48\textwidth}
        \centering
        \includegraphics[width=\textwidth]{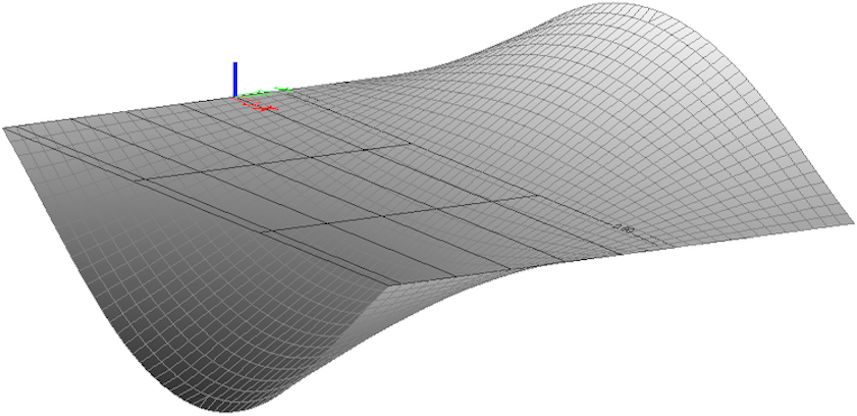}
        \caption{}    
        \label{fig:combinationmode}
    \end{subfigure}
    \hfill
    \begin{subfigure}[b]{0.48\textwidth}  
        \centering 
        \includegraphics[width=\textwidth]{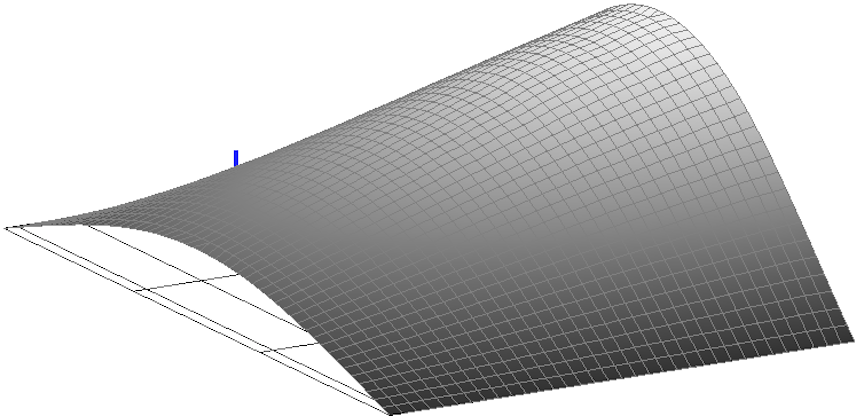}
        \caption{}    
        \label{fig:torsionalmode}
    \end{subfigure}
    \vskip\baselineskip
    \begin{subfigure}[b]{0.48\textwidth}   
        \centering 
        \includegraphics[width=\textwidth]{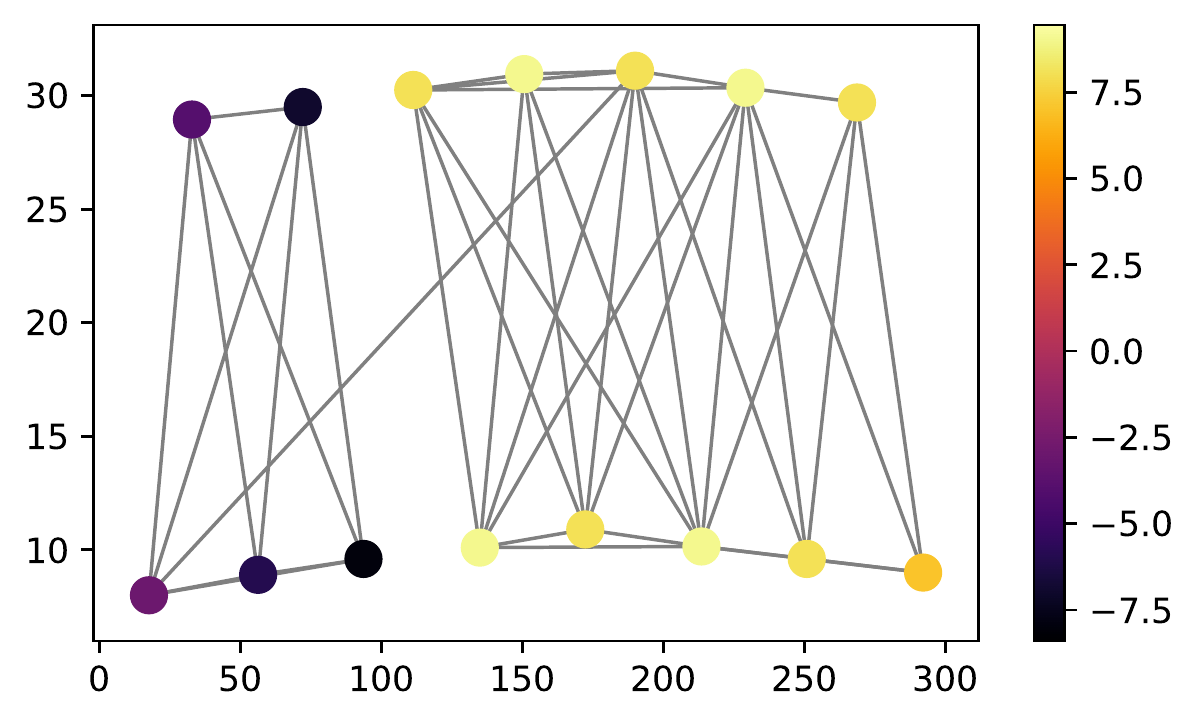}
        \caption{}    
        \label{fig:vibrationnetwork}
    \end{subfigure}
    \hfill
    \begin{subfigure}[b]{0.48\textwidth}   
        \centering 
        \includegraphics[width=\textwidth]{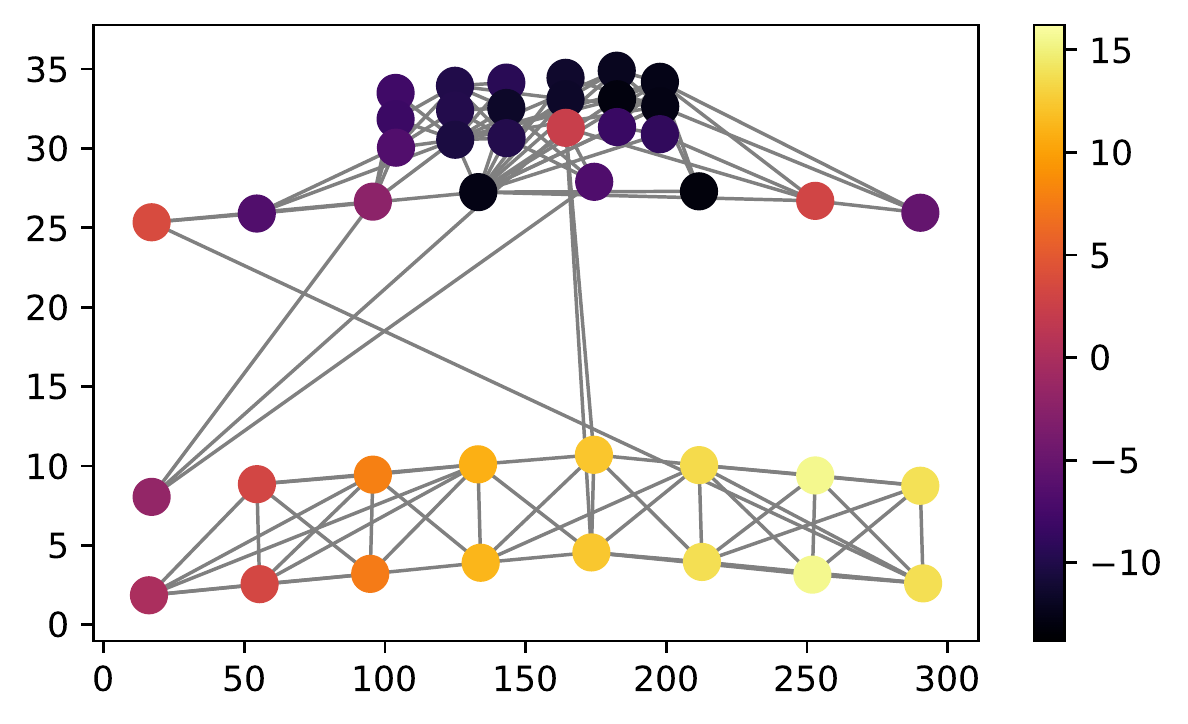}
        \caption{}    
        \label{fig:xstrainnetwork}
    \end{subfigure}
    \caption{(a) displays a FEM-based presentation of a mixture of mode shapes with the matching graph signal in (c). Figure 13 (b) presents a FEM-based mixture of torsional mode shapes that are apparent in the girders, which are as well visible in (d). Both (a) and (b) are from \citep{miao2013modal}.}
    \label{fig:modeshapeoverview}
\end{figure*}

Certain mode shapes could be observed when the graph is examined for $t$ time periods. 
For example, Figure~\ref{fig:combinationmode} shows a FEM-based mode shape, which is similar to the graph signal shown in Figure~\ref{fig:vibrationnetwork}. 
The bridge is vibrating back and forth during this event, of which Figure \ref{fig:vibrationnetwork} highlights left-sided decrease in vibration.
For example, Figure~\ref{fig:xstrainnetwork} shows a vehicle passing by on the right side of the bridge, and how the girders carry the weight and allow the other parts of the bridge to decrease in strain level. A supplementary page\footnote{\href{https://github.com/StefanBloemheuvel/GSP_Bridge}{Link} to github page} with animated GIFs is available to provide additional insights into the graph signals.

\subsection{Forecasting Strain \& Vibration}

The results of the forecast with T-GCN in terms of RMSE are visible in Table \ref{tgcn}. Overall, the T-GCN outperformed the benchmark (last observed value) in terms of RMSE by around 21\%.
It is interesting to see that the forecasting scores follow a similar pattern as the node subset sampling results. In general, the X-strain sensors are easiest to forecast, followed by combining both the X and Y-strain and lastly only Y-strain sensors. 

Considering the difference between the correlation-based graph and the DTW-based graph, not a huge disparity is found. The results in the non-filtered condition are identical, only the filtered conditions show some improvements in the Y-strain and combined sensor settings. 
However, it is recommended to use correlation over DTW in large graph applications, since calculating the DTW-based graph takes considerably longer than the correlation graph, especially if the graph is not static and thus needs to be recalculated frequently.

\begin{table}[ht]
\centering
\caption{RMSE scores for the Graph Neural Network and Benchmark baseline on forecasting strain sensor readings in the correlation \& DTW graph.}
\label{tgcn}
\begin{tabular}{@{}lccccccc@{}}
\toprule
Algorithm & \multicolumn{7}{c}{Sensor Type} \\ \cmidrule(l){2-8}
 & \multicolumn{3}{c}{Non-Filtered} & & \multicolumn{3}{c}{Filtered} \\ \cmidrule(lr){2-4} \cmidrule(lr){6-8}
 & X-Strain & Y-Strain & Combined & &  X-Strain & Y-Strain & Combined \\ \midrule
\textbf{Correlation} &      &      &      & &      &      &       \\ 
Benchmark                & 0.49 & 0.79 & 0.58 & & 0.36 & 0.59 & 0.45   \\
T-GCN                  & 0.38 & 0.61 & 0.44 & & 0.29 & 0.46 & 0.36  \\ \noalign{\vskip 3mm}
\textbf{DTW}         &      &      &      & &      &      &       \\
Benchmark                & 0.49 & 0.79 & 0.58 & & 0.34 & 0.49 & 0.37  \\
T-GCN                  & 0.38 & 0.61 & 0.44 & & 0.28 & 0.39 & 0.31  \\ \bottomrule

\end{tabular}
\end{table}

\begin{table}[ht]
\centering 
\caption{Parameter reduction when using node subset sampling on the T-GCN.}
\label{tgcnimprovements}
\begin{tabular}{@{}llll@{}}
\toprule
Situation & N parameters & Ms / epoch & RMSE \\ \midrule
All nodes & 41.170 & 10 & 0.38 \\
25\% selected & 33.346 (-19\%) & 7 (-30\%) & 0.41 (+7,9\%) \\ \bottomrule
\end{tabular}
\end{table}

\begin{figure}[ht]
\resizebox{\columnwidth}{!}{%
    \begin{tikzpicture}
    \node[inner sep=0pt] (russell) at (0,0) {\includegraphics[width=.85\textwidth]{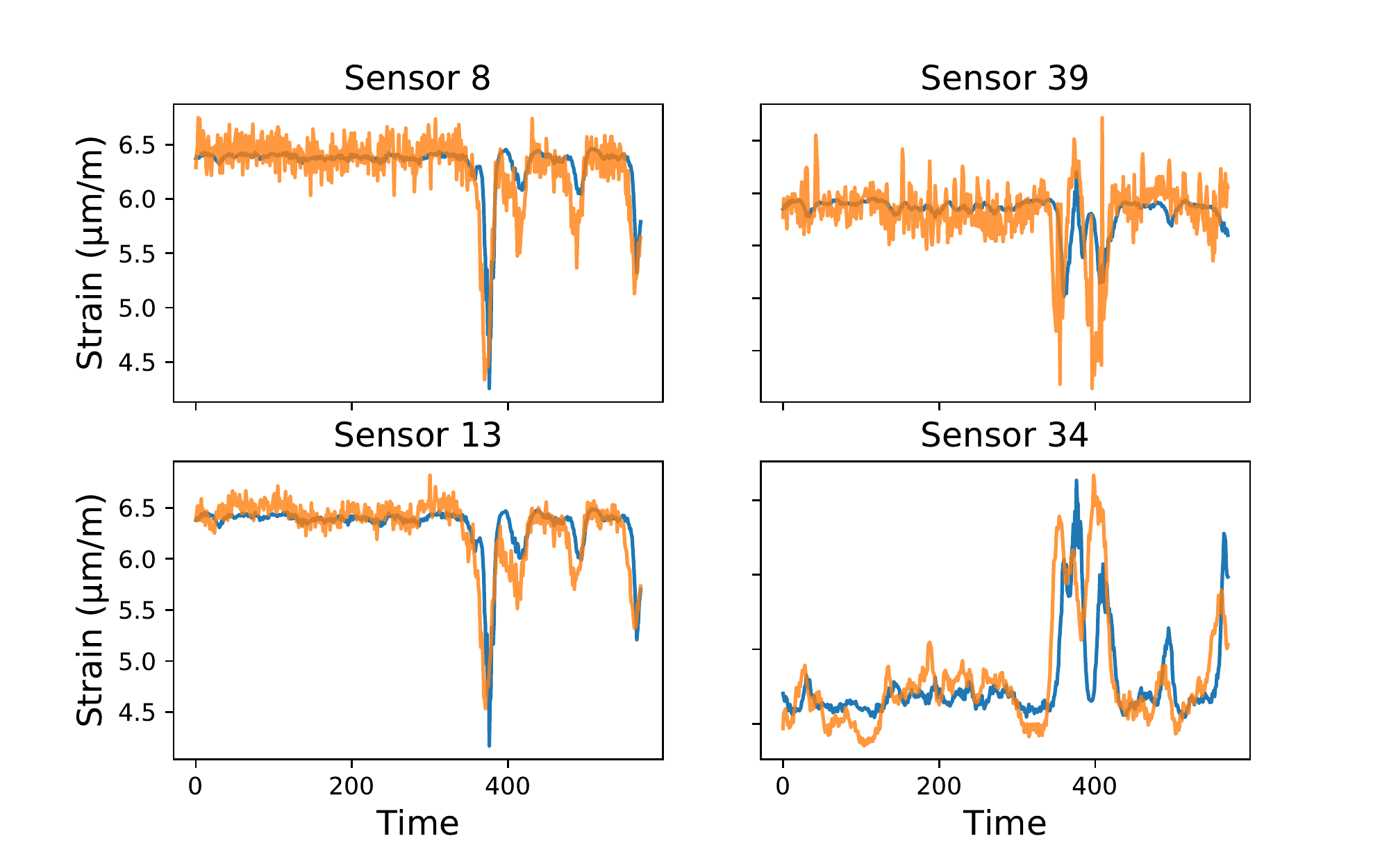}};
    \node[inner sep=0pt] (whitehead) at (7,0) {\includegraphics[width=.45\textwidth]{modeshape.pdf}};
    
    \draw[color=green!40!gray, very thick](5.01,0.63) circle (0.12); 
    \draw[color=green!40!gray, very thick](5.485,-0.55) circle (0.12); 
    \end{tikzpicture}%
}%
\caption{Re-scaled forecasting results of a set of four strain sensors and the sensor locations of sensor 34 and 39 highlighted by the green circles. The orange lines resemble the original signal, the blue lines the predicted values. Sensors 8, 13 and 39 show that the T-GCN is able to forecast the strain values, while sometimes struggling with the magnitude of events. In addition, there seems to be some delay in the forecast results of sensor 34.}\label{fig:forecaststrainplot}%
\end{figure}

The final results of the forecasting procedure are visualized in Figure \ref{fig:forecaststrainplot}. Overall, the T-GCN shows very promising results in forecasting the strain. Sensor 8 \& 13 show some similar behavioural patterns in the strain readings. The T-GCN is able to precisely forecast the strain on the bridge, however, it does struggle with predicting the magnitude of the events. Sensor 34 in Figure \ref{fig:forecaststrainplot} shows the T-GCN struggling a bit with the timing of the events that occur in the sensor. Such a delay can be due to the sub-optimal performance of the T-GCN on this specific sensor, or the fact that the sensors had to be time-aligned (see Section \ref{sec:dataset}). Lastly, sensor 39 shows a different behavioral pattern with fewer low-peaks in the strain of the bridge. 

An interesting combination between GSP and GNNs can be conducted by applying the forecasting on the subset of nodes found by GSP. In the GSP method, the set of used nodes were reduced by 75\%, leaving only 25\% selected. Such a selection boils down to a parameter reduction of 19\%, a Ms / epoch reduction of 30\% and a (tolerable) RMSE increase of 7.9\% of the selected sensors (see Table~\ref{tgcnimprovements}). Such a reduction could help combat the problem that Condition Health Monitoring systems generally collect vasts amounts of data, making analysis slow \citep{wan2018bayesian}.

\section{Conclusions}\label{sec:conclusions}
This work presented a computational framework using Graph Signal Processing and Graph Neural Networks for modeling complex sensor data and its respective analysis in the area of Structural Health Monitoring. That is, in this framework, we integrated Graph Signal Processing and Graph Neural Networks, covering more analytics oriented as well as more predictive/forecasting oriented techniques. In our experiments, we focused on a real-world complex sensor dataset in the context of structural health monitoring.
According to the results of our experiments with respect to the proposed frameworks and the respective approaches, both techniques revealed to be appropriate to work with respect to the applied real-world complex sensor data.

Our results conducted on a real-world dataset indicate that GSP is capable of choosing the most essential sensors in the \emph{Hollandse Brug}, a large bridge in the Netherlands, to derive a minimal subset of sensors from a resource-aware perspective. We also considered different strategies for network creation, investigating correlation-based and DTW-based network models. Our proposed top-down algorithm performed best of the tested alternatives in combination with the DTW-based network. With this, significant cost-reductions could be accomplished by using GSP for sensor selection in monitoring major civil infrastructures. Moreover, the sensor selection might improve the lifetime of battery-powered sensor networks, \eg by finding the two most optimal sets of sensor to turn on interchangeably.

Furthermore, we presented a method to observe (a mixture of) mode shapes, which indicate interesting events; these results could be exploited to evaluate the condition of the bridge, since the mode shapes hint to global aspects of the bridge, such as damping and stiffness. Here, our GSP approach needs fewer modeling assumptions or background knowledge in engineering (\eg compared to construction a FEM model). 

Lastly, Graph Neural Networks were used to forecast the strain values in the bridge. The T-GCN algorithm surpassed the benchmark in each condition by around 21\%. It is also interesting to note that filtering the graph signals with low-pass filters has an equal effect on reducing the forecast error as on the signal recovery using GSP. However, a possible downside of using Deep Learning is the computational complexity of such approaches. Nonetheless, the insights from the subset selection could also be used to (with a minor increase in RMSE) reduce the parameter size of the T-GCN by ${\sim}20\%$, training speed with ${\sim}30\%$ and the amount of data with 75\%.

So, in summary, in our experimentation on our real-world use case we showed that GSP enables the identification of the most important sensors, for which we investigate a set of search and optimization approaches. Furthermore, as indicated in our experiments GSP enabled the detection of specific graph signal patterns (mode shapes), capturing physical functional properties of the sensors in the applied complex network. Finally, we showed the efficacy of applying GNNs for strain prediction on this kind of data.

For future research, we intend to examine means to spot mode shapes with GSP with unsupervised techniques, \eg by adapting methods from the field of anomaly detection~\citep{Akoglu:15,AAS:17}, and also to investigate Deep Learning methods in this context. Here, particular explainable~\citep{barredo2019explainable} and interpretable~\citep{rudin2019} approaches seem interesting and relevant, \eg building on approaches combining network-based approaches with deep learning, \eg~\citep{SA:18:BNAIC,SA:21}. This also extends to further hybrid computational approaches, \eg~\citep{bellary2010hybrid,barredo2019explainable,dellermann2019hybrid}. In addition, we intend to apply according methods using GSP and GNNs on other civil infrastructures and complex systems,

In addition, investigating methods of learning the network structure in a more automated way looks promising: For this, we could consider learning the laplacian matrix, \ie constructing the graph laplacian from the data itself in a statistical or unsupervised manner could be an interesting direction for future research \citep{egilmez2016graph,dong2016learning}. This can then aid in supporting the many modeling decisions that had to be taken in order to create the graph, which are often not easy to define \citep{stankovic2020graph}. Then, it would also be interesting to compare the performance of (1) our model, (2) more semi-automatic versions, (3) and completely automatic versions for obtaining the network structure.

\section*{Acknowledgement}
We thank Dr. A.J. Knobbe for assisting with his domain knowledge that he gathered during managing the InfraWatch project. In addition, this work has been funded by the Interreg North-West Europe program (Interreg NWE), project Di-Plast - Digital Circular Economy for the Plastics Industry (NWE729).

\end{document}